# New Pruning Method Based on DenseNet Network for Image Classification

Rui-Yang Ju
Tamkang University
jryjry1094791442@gmail.com

Ting-Yu Lin
National Cheng Kung University
tonylin0413@gmail.com

Jen-Shiun Chiang
Tamkang University
chiang@mail.tku.edu.tw

*Abstract* — Deep neural networks have made significant progress in the field of computer vision. Recent works have shown that depth, width and shortcut connections of the neural network architectures play a crucial role in their performance. As one of the most advanced neural network architectures, DenseNet, which achieves excellent convergence speed through dense connections. However, it still has obvious shortcomings in the use of memory. In this paper, we introduce two new pruning methods using threshold, which refers to the concept of threshold voltage in MOSFET. Now we have implemented one of the pruning methods. This work uses this method to connect blocks of different depths in different ways to reduce memory usage. We name the proposed network ThresholdNet, evaluate it and other different networks on two datasets (CIFAR-10 and STL-10). Experiments show that the proposed method is 60% faster than DenseNet, 20% faster and 10% lower error rate than HarDNet.

*Keywords* — *Convolutional neural network, Computer vision, Deep learning, Threshold, Pruning*

## I. INTRODUCTION

Convolutional neural networks (CNN) are very popular in the applications of computer vision (CV), such as image classification, object detection, and semantic segmentation [17-21]. With the enhancement of hardware computing power, deep neural networks (DNN) such as DenseNet [3], GoogLeNet [12], and SqueezeNet [13] can be completed in a reasonable time. Among these networks, one of the most advanced networks, DenseNet, can achieve faster convergence speed than ResNet [4], but its shortcomings are also obvious. On one hand, DenseNet needs to access memory frequently and is therefore more hardware-constrained. Besides, too deep network architecture may cause overfitting problems, resulting in lower accuracy. So how to increase computation efficiency and reduce the power consumption for neural networks become a critical issue. The latest DenseNet pruning work, HarDNet [5], uses harmonic dense connections instead of dense connections, which greatly reduces the number of connections between layers and reduces the amount of memory. However, the accuracy of HarDNet suffers to a certain extent, because its indirectness ignores the connections of odd layers.

The contribution of this paper is to use a threshold to segment the convolutional layer of different depths. For convolutional layers with smaller depths, we can use dense connections or less pruning dense connections; for larger depths convolutional layers, we prefer to use harmonic dense connections. Therefore, the improved neural network can greatly improve training speed while minimizing the accuracy impairment.

## II. RELATED WORK

### A. Convolutional neural networks

Since 2012, as a new model architecture, neural networks have achieved excellent results in various fields of CV. AlexNet [1] is composed of 8 convolutional layers and won the ILSVRC-2012 image classification championship. It reduced the error rate on ImageNet dataset from 25.8% to 16.4%. In 2014, VGG-Net [2] was introduced to easily extend the depth of neural networks to 20 levels, and the accuracy of image classification was greatly improved.

As the networks go deeper, simply stacking layers would degrade its performance. In order to solve this problem, ResNet learns residual function $H(x) - x$ to replace target function $H(x)$ directly. ResNet can extend the depth to more than 100 layers, and the performance is further improved. On this basis, ResNeXt [6] presented in CVPR 2017 combines the architecture of Inception [7-10] (Split-transform-merge) to improve the accuracy without increasing the number of parameters.

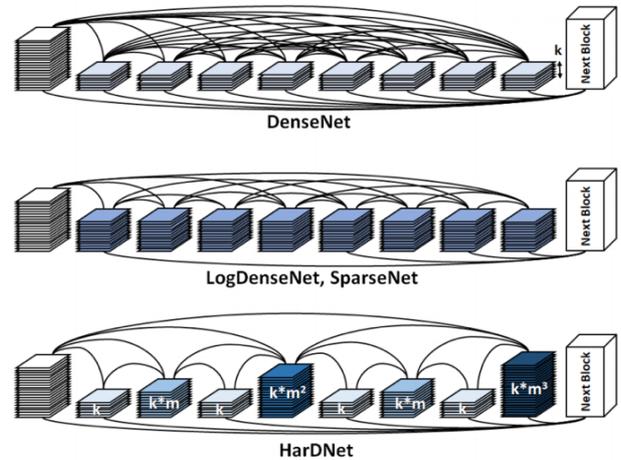

Fig. 1. Architectures of DenseNet, LogDenseNet, SparseNet, HarDNet. [5]

### B. Densely connected convolutional networks

The CVPR 2017 best paper DenseNet, based on the design concept of ResNet, proposes to realize the feature reuse through the connection of features on the channel, and densely connects all the front layers with dense connections. DenseNet is composed of multiple dense blocks, and each block is composed of several convolutional layers. Each convolutional layer generates growth rate (*k*) features, where *k* refers to the number of feature maps generated by the $H(x)$ function. As shown by DenseNet in Fig. 1, for an *L* layer network, DenseNet contains a total of $(L(L + 1))/2$ connections.

## C. Harmonic densely connected networks

With the improvement of computing power and the expansion of datasets, scholars have started to train more complex networks. Therefore, how to reduce power consumption while ensuring computing efficiency has become a critical issue. In this case, HarDNet replaces the dense connections in DenseNet with harmonic dense connections. As shown by HarDNet in Fig. 1, the network is like multiple harmonics, and the dense connections are pruned in a regular manner to effectively reduce the power consumption of the network architecture.

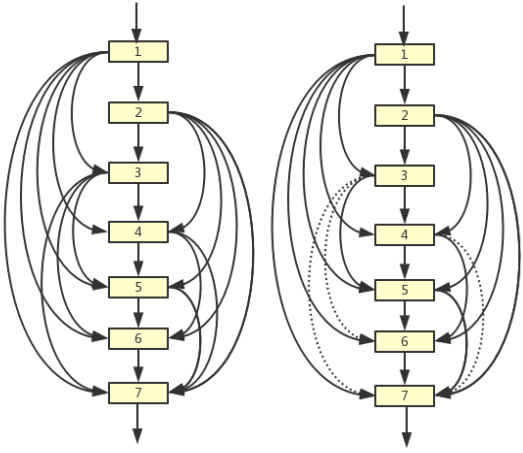

Fig. 2. The left figure is a DenseNet, and the inputs to the layers are from all the previous layers; the right figure is a SparseNet, and the dotted lines are dropped connections. The inputs to the layers are from at most two previous layers. [14]

## D. Sparse densely connected networks

In recent years, the research on DenseNet pruning has not stopped. Liu *et al*. [14] proposed another method to sparse DenseNet, who reduce the connections of the original L-layer DenseNet from $O(L^2)$ to $O(L)$. As shown in Fig. 2, the authors delete the connections in the middle layer based on dense connections in DenseNet and only keeps the farthest and nearest connections. The formula is as follows in (1):

$$x_i = H\left(\left[x_0, x_1, \cdots, x_{\frac{n}{2}}, x_{i-\frac{n}{2}}, \cdots, x_{i-1}\right]\right) \quad (1)$$

where *n* is the number of connections to be retained.

## III. PROPOSED METHODS

This work proposes the concept of threshold, which makes the architecture of the deep neural networks more diversified. In addition, we propose a new network architecture based on DenseNet, HarDNet and SparseNet. We propose two pruning methods. The first is to set Threshold between block and block, and the second is to set Threshold between Layer and Layer. The formulas for both ways can be summarized as (2) and (3)

$$i \le Threshold, x_i = H([x_0, x_1, \cdots, x_{i-1}]) \quad (2)$$

$$i > Threshold, \begin{cases} if\ i\ \%\ 2 = 1, x_i = H([x_{i-1}]) \\ else, x_i = H([x_0, \cdots, x_{i-2^m}]) \end{cases}$$

$$i \le Threshold, x_i = H\left(\left[x_0, x_1, \cdots, x_{\frac{n}{4}}, x_{i-\frac{n}{4}}, \cdots, x_{i-1}\right]\right) \quad (3)$$

$$i > Threshold, \begin{cases} if\ i\ \%\ 2 = 1, x_i = H([x_{i-1}]) \\ else, x_i = H([x_0, \cdots, x_{i-2^m}]) \end{cases}$$

## A. Threshold

A threshold method is proposed in which the threshold block becomes sparse to vary degrees with more convolutional layers. The size of the threshold value is positively related to the complexity of the neural network. Then we set the threshold to different values and obtain the corresponding neural network architecture according to changing the magnitude of the threshold. As shown in (2), we compare the *i*-th layer in the threshold block with the threshold value. If the value of *i* is smaller than the threshold value, then the dense connection mode will be maintained. If the value of *i* is larger than the threshold value, the original connection method is replaced with a harmonic dense connection.

TABLE I. The connection when Threshold = 4 and Layer = 16

| Layer | Connection | Layer | Connection |
|---|---|---|---|
| *n* = 1 | 0 | *n* = 9 | 8 |
| *n* = 2 | 0-1 | *n* = 10 | 2-6-8-9 |
| *n* = 3 | 0-1-2 | *n* = 11 | 10 |
| *n* = 4 | 0-1-2-3 | *n* = 12 | 4-8-10-11 |
| *n* = 5 | 4 | *n* = 13 | 12 |
| *n* = 6 | 4-5 | *n* = 14 | 6-10-12-13 |
| *n* = 7 | 6 | *n* = 15 | 14 |
| *n* = 8 | 0-4-6-7 | *n* = 16 | 0-8-12-14-15 |

In addition, we propose a new connection method, as shown in (3). If the value of *i* is less than the threshold value, we sparse the original dense connections, and retain the connections between the nearest and farthest layers. In the formula, *n* is the value of threshold. It deletes the connection between (*n*/4) and (*i*–*n*/4), and it can reduce a total of (*i*–*n*/2) connections. If the value of the threshold is greater, the number of sparse connections will be less, and the complexity of the neural network architecture will be higher, ensuring that the size of the threshold is positively correlated with the complexity of the neural network.

TABLE II. The connection when Threshold = 8 and Layer = 16

| Layer | Connection | Layer | Connection |
|---|---|---|---|
| *n* = 1 | 0 | *n* = 9 | 8 |
| *n* = 2 | 0-1 | *n* = 10 | 2-6-8-9 |
| *n* = 3 | 0-1-2 | *n* = 11 | 10 |
| *n* = 4 | 0-1-2-3 | *n* = 12 | 4-8-10-11 |
| *n* = 5 | 0-1-2-3-4 | *n* = 13 | 12 |
| *n* = 6 | 0-1-2-4-5 | *n* = 14 | 6-10-12-13 |
| *n* = 7 | 0-1-2-5-6 | *n* = 15 | 14 |
| *n* = 8 | 0-1-2-6-7 | *n* = 16 | 0-8-12-14-15 |

## B. First purning method

Our first pruning method is to set the thresholds to 4, 8 and 12 respectively for sparse processing, using the block with layer = 16 as the sparse object, and use sparse dense connection and harmonic dense connection to connect. As shown in TABLE I, threshold = 4 has the greatest degree of pruning, with a total of 38 connections; as shown in TABLE II, there are 51 connections when threshold = 8; as shown in TABLE III, threshold = 12 has the smallest degree of pruning, with a total of 74 connections.

As shown in Fig. 3, for layers smaller than the threshold we will output all their channels, and for layers larger than the threshold we will only output odd-numbered layers. In order to increase the weight of the key layers, we do this by increasing their channels. The initial growth rate of the $i$-th layer is set to $k$, and let the number of channels in the even-numbered layer be $k \times m$, where $m$ is the compression factor. If the input layer is 0 and $m = 2$, the channel ratio of each layer is 1 : 1, so set $m$ to a value less than 2 to compress the number of channels in the even-numbered layers.

TABLE III. The connection when Threshold = 12 and Layer = 16

| Layer | Connection | Layer | Connection |
|---|---|---|---|
| $n = 1$ | 0 | $n = 9$ | 0-1-2-3-6-7-8 |
| $n = 2$ | 0-1 | $n = 10$ | 0-1-2-3-7-8-9 |
| $n = 3$ | 0-1-2 | $n = 11$ | 0-1-2-3-8-9-10 |
| $n = 4$ | 0-1-2-3 | $n = 12$ | 0-1-2-3-9-10-11 |
| $n = 5$ | 0-1-2-3-4 | $n = 13$ | 12 |
| $n = 6$ | 0-1-2-3-4-5 | $n = 14$ | 6-10-12-13 |
| $n = 7$ | 0-1-2-3-4-5-6 | $n = 15$ | 14 |
| $n = 8$ | 0-1-2-3-5-6-7 | $n = 16$ | 0-8-12-14-15 |

The harmonic dense connection proposed by HarDNet greatly prunes the dense connection of DenseNet, but it neglects the connection of the odd-numbered layers. All odd-numbered layers are only connected to the upper layer, which greatly affects the accuracy of the neural network. We find that the number of connections between the smaller $i$ layers in the threshold block is not large enough to require extensive pruning, while the number of connections between the larger $i$ layers is large enough to use harmonic dense connections.

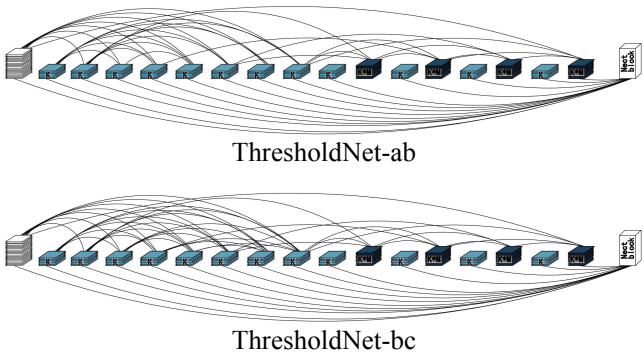

Fig. 3. (a) ThresholdNet-ab, (b) ThresholdNet-bc.

In this pruning method, we propose two network architectures. As shown in Fig. 3, Fig. 3(a) is ThresholdNet-ab; here we set threshold = 8, use dense connections for layers less than the threshold value, and use harmonics for layers greater than the threshold value. On the other hand Fig. 3(b) is ThresholdNet-bc; here we set threshold = 8, use sparse dense connection for layers less than the threshold value, and a harmonic dense connection for layers greater than the threshold value.

## C. Second purning method

Our second pruning method also uses the Threshold concept of distinction, but is different from the previous method. The second method uses the distinction of blocks to connect different blocks differently. As shown in TABLE IV, we change the original block of DenseNet (containing 24 convolutions) to two blocks (containing 12 convolutions and 16 convolutions, respectively). We use different connections for these two blocks. The former block uses dense connections, and the latter block uses harmonic dense connections.

This method is different from the first type of pruning. It makes the connection mode within each block consistently and ensures that different blocks have different connection modes. The concept of Threshold is also used, and pruning can still be carried out reasonably.

## D. Detailed architecture design

As shown in TABLE V, we set the input image size to 224 × 224, and propose a new threshold block, followed by a 1×1 convolutional layer, and no pooling layer is added between the blocks that are segmented by the threshold. .

We have implemented the second pruning method and proposed ThresholdNet_v1. For batch normalization (BN) [22] and then rectified linear unit (ReLU) [23], our proposed network replaces the BN-ReLU-Conv sequence with the DenseNet standard sequence of Conv-BN-ReLU. In addition, we removed the 7 × 7 Conv used by DenseNet instead of using two 3 × 3 Convs as the initial convolutional layer. On this basis, we improved the depth of the network architecture and proposed a more complex ThresholdNet_v2.

In order to compare the improved architecture and the memory of DenseNet in detail, we follow the growth rate = 32 proposed by DenseNet121. In addition, we set the reduction of the output channel for layers smaller and larger than the threshold differently, with the reduction set to 0.5 for the densely connected part and 0.1 for the harmonic dense part.

TABLE IV. Architecture comparison

| DenseNet | ThresholdNet | |
|---|---|---|
| Block | Block | |
| 24 | 12 | 16 |
| $x_i = H([x_0, x_1, \cdots, x_{i-1}])$ | $x_i = H([x_0, x_1, \cdots, x_{i-1}])$ | $\begin{cases} if\ i\ \%\ 2 = 1, x_i = H([x_{i-1}]) \\ else\ x_i = H([x_0, \cdots, x_{i-2^m}]) \end{cases}$ |
| 32 | 16 | 16 |
| $x_i = H([x_0, x_1, \cdots, x_{i-1}])$ | $x_i = H([x_0, x_1, \cdots, x_{i-1}])$ | $\begin{cases} if\ i\ \%\ 2 = 1, x_i = H([x_{i-1}]) \\ else, x_i = H([x_0, \cdots, x_{i-2^m}]) \end{cases}$ |

TABLE V. Detailed implementation parameters of the proposed architecture

| Layers | Output size | ThresholdNet_v1 | ThresholdNet_v2 |
|---|---|---|---|
| Convolution | 112 × 112 | 3 × 3 conv | |
| | 112 × 112 | 3 × 3 conv | |
| Pooling | 56 × 56 | 3 × 3 max pool | |
| Threshold Block (1) | 56 × 56 | [ 3 × 3 conv] × 6 | [ 3 × 3 conv] × 6 |
| Transition Layer (1) | 56 × 56 | 1 × 1 conv | |
| | 28 × 28 | 2 × 2 max pool | |
| Threshold Block (2) | 28 × 28 | [ 3 × 3 conv] × 8 | [ 3 × 3 conv] × 12 |
| Transition Layer (2) | 28 × 28 | 1 × 1 conv | |
| | 14 × 14 | 2 × 2 max pool | |
| Threshold Block (3) | 14 × 14 | [ 3 × 3 conv] × 12 | [ 3 × 3 conv] × 16 |
| Transition Layer (3) | 14 × 14 | 1 × 1 conv | |
| Threshold Block (4) | 14 × 14 | [ 3 × 3 conv] × 16 | [ 3 × 3 conv] × 16 |
| Transition Layer (4) | 14 × 14 | 1 × 1 conv | |
| | 7 × 7 | 2 × 2 max pool | |
| Threshold Block (5) | 7 × 7 | [ 3 × 3 conv] × 4 | [ 3 × 3 conv] × 4 |
| Classification Layer | 1 × 1 | 1 × 1 average pool | |
| | | 1000D fully-connected, softmax | |

## IV. EXPERIMENT RESULTS

We use the second pruning method to complete the network architecture ThresholdNet_v1 and ThresholdNet_v2 for training on the CIFAR-10 and STL-10 datasets. CIFAR-10 [15] is a 32 × 32 size RGB color image with 50,000 training data and 10,000 test data. STL-10 [16] is a collection of 10 types of object pictures, and each type has 1,300 pictures, 500 of which are used for training and 800 are used for testing, and each picture has a resolution of 96 × 96.

We improve the original dense unit proposed by DenseNet, add the dense unit with Layer 24 in DenseNet121 to the concept of threshold, and split it into 8 and 16 threshold units. The original block has connection methods, and the improved threshold block has different connection methods for different layers.

It can be seen from TABLE VI that on CIFAR-10, although DenseNet and ResNet have good accuracy, their calculation time on GPU is too long. HarDNet can cut the time by half, but its accuracy needs to be improved. ThresholdNet has increased the speed while ensuring that the accuracy rate is not affected, and finally achieved a 20% faster and 10% lower error rate than HarDNet. In addition, on STL-10, as shown in TABLE VII, ThresholdNet can still have the same accuracy rate and is 10% faster than HarDNet.

In order to analyze the impact of threshold on pruning in detail, DenseNet, ResNet, ThresholdNet, ResNeXt and wide residual networks (WRN) [11] were tested in parameters, memory and floating-point operations per second (FLOPS). From the detailed data in TABLE VIII, we can see that compared with HarDNet68, our network effectively reduces Params and MAdds.

TABLE VI. CIFAR-10 Experimental Results

| | Error (%) | Step time (ms) | Total time (s) |
|---|---|---|---|
| DenseNet121 | 7.35 | 119 | 59.13 |
| ResNet50 | 9.89 | 102 | 50.37 |
| ThresholdNet_v1 | 13.66 | 47 | 24.21 |
| ThresholdNet_v2 | 13.31 | 54 | 27.25 |
| HarDNet68 | 14.66 | 57 | 27.98 |
| ResNeXt50_32x4d | 16.96 | 52 | 24.81 |
| Wide_ResNet50_2 | 17.31 | 55 | 27.86 |

Test results for CIFAR-10 datasets, in which Step Time and Total Time are measured on Nvidia GTX3080Ti with Pytorch 1.1.0. Step Time is time for each batch, and Total Time is time for each epoch.

TABLE VII. STL-10 Experimental Results

| | Error (%) | Step time (ms) | Total time (s) |
|---|---|---|---|
| DenseNet121 | 25.53 | 175 | 8.77 |
| ResNet18 | 23.62 | 186 | 9.61 |
| ThresholdNet_v1 | 24.95 | 87 | 4.41 |
| ThresholdNet_v2 | 24.30 | 115 | 5.39 |
| HarDNet68 | 24.98 | 113 | 4.79 |
| ResNeXt50_32x4d | 36.35 | 110 | 5.30 |
| Wide_ResNet50_2 | 32.23 | 132 | 6.39 |

Test results for STL-10 datasets, in which Step Time and Total Time are measured on Nvidia GTX 3080Ti with Pytorch 1.1.0. Step Time is time for each batch, and Total Time is time for each epoch.

TABLE VIII. Detailed Data

|  | Block_Depth | #Params | #MAdds | #FLOPS |
|---|---|---|---|---|
| DenseNet121 | 6, 12, 24, 16 | 7.97M | 2.88G | 2.88G |
| DenseNet169 | 6, 12, 32, 32 | 14.15M | 3.42G | 3.42G |
| DenseNet201 | 6, 12, 48, 32 | 20.01M | 8.70G | 4.37G |
| DenseNet161 | 6, 12, 36, 24 | 28.68M | 15.6G | 7.82G |
| ResNet50 | 3, 4, 6, 3 | 25.55M | 8.22G | 4.12G |
| ResNet101 | 3, 4, 23, 3 | 44.55M | 15.66G | 7.84G |
| HarDNet68 | 8, 16, 16, 16, 4 | 17.57M | 8.51G | 4.26G |
| ThresholdNet_v1 | 6, 8, 12, 16, 4 | 15.32M | 6.90G | 3.46G |
| ThresholdNet_v2 | 6, 12, 16, 16, 4 | 17.14M | 8.12G | 4.07G |
| ResNeXt50_32x4d | 3, 4, 6, 3 | 25.03M | 8.51G | 4.27G |
| ResNeXt101_32x8d | 3, 4, 23, 3 | 88.79M | 32.93 | 16.49G |
| Wide_ResNet50_2 | 3, 4, 6, 3 | 68.88M | 22.85G | 11.43G |
| Wide_ResNet101_2 | 3, 4, 23, 3 | 126.89M | 45.58G | 22.81G |

## V. CONCLUSION

Before this paper was put forward, DenseNet has always the problem of a large amount of memory used, which cannot be implemented well on the hardware, which hindered the development of related researches. The researches on DenseNet pruning have not stopped. Most of them only use a single sparse method, discarding the accuracy of the neural network to reduce the amount of memory.

The threshold approach proposed in this paper is pruned. The experimental results show that by choosing different threshold values, we can choose the weight between the accuracy and the amount of memory by ourselves, and complete the work of complex neural network pruning more reasonably. The problem of hardware implementation has promoted the research and development of CNN in hardware.

## REFERENCES

[1] A. Krizhevsky, I. Sutskever, and G. Hinton, "ImageNet classification with deep convolutional neural networks," *Advances in Neural Information Processing Systems*, vol. 25, pp. 1097-1105, 2012.

[2] K. Simonyan and A. Zisserman, "Very deep convolutional networks for large-scale image recognition," 2014, *arXiv:1409.1556*.

[3] G. Huang, Z. Liu, K.Q. Weinberger, and L. van der Maaten, "Densely connected convolutional networks", *IEEE Conference on Computer Vision and Pattern Recognition*, Jul. 2017, pp. 2261-2269.

[4] K. He, X. Zhang, S. Ren, and J. Sun, "Deep residual learning for image recognition", *IEEE Conference on Computer Vision and Pattern Recognition*, Jun. 2016, pp. 770-778.

[5] P. Chao, C. Kao, Y. Ruan, C. Huang, and Y. Lin, "HarDNet: A low memory traffic network," *International Conference on Computer Vision*, Oct. 2019, pp. 3551-3560.

[6] S. Xie, R. Girshick, P. Dollár, Z. Tu, and K. He, "Aggregated residual transformations for deep neural networks," *IEEE Conference on Computer Vision and Pattern Recognition*, Jul. 2017, pp. 5987-5995.

[7] C. Szegedy, W. Liu, Y. Jia, P. Sermanet, S. Reed, D. Anguelov, D. Erhan, V. Vanhoucke, and A. Rabinovich, "Going deeper with convolutions," 2014, *arXiv:1409.4842*.

[8] S. Ioffe and C. Szegedy, "Batch normalization: Accelerating deep network training by reducing internal covariate shift," 2015, *arXiv:1502.03167*.

[9] C. Szegedy, V. Vanhoucke, S. Ioffe, J. Shlens, and Z. Wojna, "Rethinking the inception architecture for computer vision," 2015, *arXiv:1512.00567*.

[10] C. Szegedy, S. Ioffe, V. Vanhoucke, and A. Alemi, "Inception-v4, Inception-ResNet and the impact of residual connections on learning," 2016, *arXiv:1602.07261*.

[11] S. Zagoruyko and N. Komodakis, "Wide residual networks," 2016, *arXiv:1605.07146*.

[12] C. Szegedy, W. Liu, Y. Jia, P. Sermanet, S. Reed, D. Anguelov, D. Erhan, V. Vanhoucke, and A. Rabinovich, "Going deeper with convolutions", *IEEE Conference on Computer Vision and Pattern Recognition*, Jun. 2015, pp. 1063-6919.

[13] F. N. Iandola, S. Han, M. W. Moskewicz, K. Ashraf, W. J. Dally, and K. Keutzer, "Squeezenet: Alexnet-level accuracy with 50x fewer parameters and <0.5MB model size", 2016, *arXiv:1602.07360*.

[14] W. Liu and K. Zeng, "SparseNet: A sparse DenseNet for image classification", 2016, *arXiv:1804.05340*.

[15] A. Krizhevsky and G. Hinton, "Learning multiple layers of features from tiny images", *Tech Report*, 2009.

[16] A. Coates, H. Lee, and A. Y. Ng, "An analysis of single-layer networks in unsupervised feature learning", *International Conference on Artificial Intelligence and Statistics*, Apr. 2011, pp. 215-223.

[17] R. Girshick, "Fast R-CNN," *International Conference on Computer Vision*, Dec. 2015, pp. 1440-1448.

[18] E. Shelhamer, J. Long, and T. Darrell, "Fully convolutional networks for semantic segmentation," *IEEE Transactions on Pattern Analysis and Machine Intelligence*, vol. 39, no. 4, pp. 640-651, 2017.

[19] B. Hariharan, P. Arbeláez, R. Girshick, and J. Malik, "Hypercolumns for object segmentation and fine-grained localization", *IEEE Conference on Computer Vision and Pattern Recognition*, Dec. 2015, pp. 447-456.

[20] P. Sermanet, K. Kavukcuoglu, S. Chintala, and Y. LeCun, "Pedestrian detection with unsupervised multi-stage feature learning", *IEEE Conference on Computer Vision and Pattern Recognition*, Dec. 2013, pp. 3626-3633.

[21] S. Yang and D. Ramanan, "Multi-scale recognition with DAG-CNNs", *IEEE Conference on Computer Vision and Pattern Recognition*, Dec. 2015, pp. 1215-1223.

[22] S. Ioffe and C. Szegedy, "Batch normalization: accelerating deep network training by reducing internal covariate shift", *International Conference on Machine Learning*, Jul. 2015, pp. 448-456.

[23] X. Glorot, A. Bordes, and Y. Bengio, "Deep sparse rectifier neural networks", *International Conference on Artificial Intelligence and Statistics*, Apr. 2011, pp. 315-323.